\documentclass[11pt,a4paper]{article}
\usepackage[hyperref]{acl2018}
\PassOptionsToPackage{breaklinks}{hyperref}
\usepackage{times}
\usepackage{graphicx}
\usepackage{latexsym}

\usepackage{todonotes}
\usepackage{booktabs}

\usepackage{algorithm}
\usepackage[noend]{algpseudocode}
\usepackage{amsmath}
\usepackage{amsfonts}
\usepackage{stmaryrd}
\usepackage{hyperref}
\usepackage[T1]{fontenc}
\usepackage{url}

\usepackage{xcolor}

\aclfinalcopy 

\title{Generating High-Quality Surface Realizations Using Data Augmentation and Factored Sequence Models}

\author{Henry Elder \\
  ADAPT Centre, \\
  Dublin City University, Ireland \\
  {\tt henry.elder@adaptcentre.ie} \\\And
  Chris Hokamp \\
  Aylien Ltd. \\
  Dublin, Ireland \\
  {\tt chris@aylien.com} \\}

\date{}

\begin{document}
\maketitle
\begin{abstract}
This work presents a new state of the art in reconstruction of surface realizations from obfuscated text. We identify the lack of sufficient training data as the major obstacle to training high-performing models, and solve this issue by generating large amounts of synthetic training data. We also propose preprocessing techniques which make the structure contained in the input features more accessible to sequence models. Our models were ranked first on all evaluation metrics in the English portion  of the 2018 Surface Realization shared task. 


\end{abstract}

\section{Introduction}






Contextualized Natural Language Generation (NLG) is a long-standing goal of Natural Language Processing (NLP) research. The task of generating text, conditioned on knowledge about the world, is applicable to almost any domain. However, despite recent advances in specific domains, NLG models still produce relatively low quality outputs in many settings. Representing the context in a consistent manner is still a challenge: how can we condition output on a stateful structure such as a graph or a tree?

Several shared tasks have recently explored NLG from inputs with graph-like structures; RDF triples \citep{Colin2016TheData}, dialogue act-based meaning representations \citep{Novikova2017TheGeneration} and abstract meaning representations \citep{May2017}. In each of these challenges, the input has structure beyond simple linear sequences; however, to date, the top results in these tasks have consistently been achieved using relatively standard sequence-to-sequence models. 

The \textbf{surface realization} task is a conceptually simple challenge: given shuffled input, where tokens are represented by their lemmas, parts of speech, and dependency features, can we train a model to reconstruct the original text? A model that performs well at this task is likely to be a good starting point for solving more complex tasks, such as NLG from Resource Description Framework (RDF) graphs or Abstract Meaning Representation (AMR) structures. In addition, training data for the surface realization task can also be generated in a fully-automated manner.

In this work, we show that training dataset size may be the major obstacle preventing current sequence-to-sequence models from doing well at NLG from structured inputs. Although inputting the structures themselves is theoretically appealing \cite{Tai2015ImprovedNetworks}, in many domains it may be enough to use sequential inputs by flattening structures, and providing structural information via input factors, as long as the training dataset is sufficiently large. By augmenting training data using a large corpus of unannotated data, we obtain a new state of the art in the surface realization task using off-the-shelf sequence to sequence models. 

In addition, we show that information about the output word order, implicitly available from parse features, provides essential information about the word order of correct output sequences, confirming that structural information cannot be discarded without a large drop in performance.

The main contributions of this work are:

\begin{enumerate}
  \item We show how training datasets can be augmented with synthetic data
  \item We apply preprocessing steps to simplify the universal dependency structures, making the structure more explicit
  \item We evaluate pointer models for the surface realization task 
\end{enumerate}

\begin{table*}[ht!]
\centering
\begin{tabular}{p{0.15\linewidth}p{0.45\linewidth}p{0.15\linewidth}p{0.15\linewidth}}
\toprule
 \textsc{Feature} & \textsc{Description} & \textsc{Vocabulary Size} & \textsc{Embedding Size}  \\ \midrule
 lemma & the lemma of the surface word & 30004 & 300 \\
 XPOS & the English part-of-speech label & 53 & 16 \\
 position & the position in the sequence & 103 & 25 \\
 UPOS & the universal part-of-speech label & 20 & 8 \\
 head position & the position of the head word according to the dependency parser & 100 & 25 \\
 deprel & the dependency relation label according to the dependency parser & 51 & 15 \\
\bottomrule
\end{tabular}
\caption{The features used in the factored models, along with the number of possible values the feature may take, and the respective embedding size.}
\label{tab:features}
\end{table*}
\section{The Surface Realization Shared Task}

In the \textbf{shallow} track of the 2018 surface realization (SR) shared task, inputs consist of tokens from a universal dependency (UD) tree provided in the form of lemmas. The original order of the sequence is obfuscated by random shuffling\footnote{The task organizers also introduced a \textbf{deep} task, but since ours was the only submission to the deep task, we save our discussion of this task for future work.}.

Models are evaluated on their ability to reconstruct the original, unshuffled input which generated the features. In order to do this, models must make use of structural information in order to reorder the tokens correctly as well as part-of-speech and/or dependency parse labels in order to restore the correct surface realization of lemmas. Note that we focus upon the English sub-task, where word order is critical because of the typologically analytic nature of English, however, for other languages, restoring word order may be less important, while deriving surface realizations from lemmas may be much more challenging.



\section{Datasets}

\subsection{Augmenting Training with Synthetic Datasets}

To augment the SR training data, we used sentences from the WikiText corpus \citep{Merity2016PointerModels}. Each of these sentences was parsed using UDPipe \cite{udpipe:2017} to obtain the same features provided by the SR organizers. We then filtered this data, keeping only sentences with at least 95\% vocabulary overlap with the in-domain SR training data. Note that the input vocabulary for this task is word lemmas, so at least 95\% of the tokens in each instance in our additional training data are lemmas which are also found in the in-domain data. The order of tokens in each instance of this additional dataset is then randomly shuffled to simulate the random input order in the SR data. 

We thus obtain 642,960 additional training instances, which are added to the 12,375 instances supplied by the SR shared task organizers.

\begin{table*}[ht!]
\centering
\begin{tabular}{p{0.1\linewidth}p{0.15\linewidth}p{0.15\linewidth}p{0.1\linewidth}p{0.2\linewidth}p{0.1\linewidth}p{0.15\linewidth}}
\toprule
 \textsc{position} & \textsc{lemma} & \textsc{XPOS} & \textsc{UPOS} & \textsc{head position} & \textsc{deprel} \\ \midrule
 1 & learn & VERB & VB & 2 & acl \\
 2 & lot & NOUN & NN & 4 & nsubj \\
 3 & there & PRON & EX & 4 & expl \\
 4 & be & VERB & VBZ & 0 & root \\
 5 & about & ADP & IN & 8 & case \\
 6 & a & DET & DT & 2 & det \\
 7 & . & PUNCT & . & 4 & punct \\
 8 & Chernobyl & PROPN & NNP & 1 & obl \\
 9 & to & PART & TO & 1 & mark \\
\bottomrule
\end{tabular}
\caption{An example from the training data, containing all features we use as input factors.}
\label{tab:input-data-example}
\end{table*}

\section{Features}

\subsection{Leveraging Structured Features}

Because we have the dependency parse features for each input, some information about word order is implicitly available from the parse information, but discovering the structural relationship between the dependency parse features and the order of words in the output sequence is likely to be challenging for our sequence to sequence model. Therefore, we construct the original parse tree from the dependency features, and perform a depth-first search to sort and reorder the lemmas. This is similar to the linearization step performed by \citet{Konstas2017NeuralGeneration}, the main difference being we randomly choose between child nodes instead of using a predetermined order based on edge types.


In order to further augment the available context, we experiment with adding potential delemmatized forms for each input lemma. The possible forms for each lemma were found by creating a map from $ (\mathbf{lemma}, \mathbf{xpos}) \rightarrow \mathbf{form}$, using the WikiText dataset. For each input lemma and xpos, we then check for the pair in the map -- if it exists, the corresponding form is appended to the sequence. This makes forms available to the pointer model for copying. 

For some lemma, xpos pairs there are multiple potential forms. When this occurs we add all potential forms to the input sequence. The mapping was found to cover 98.9\% of cases in the development set. 

\subsection{Factored Inputs}

Factored models were introduced by Alexandrescu et al. \shortcite{Alexandrescu:2006:NLM} as a means of including additional features beyond word tokens into neural language models. The key idea is to create a separate embedding representation for each feature type, and to concatenate the embeddings for each  input token to create its dense representation. Sennrich et al. \shortcite{SennrichH16:factors} showed that this technique is quite effective for neural machine translation, and some recent work, such as Hokamp \shortcite{Hokamp:2017} has successfully applied this technique to related sequence generation tasks. 

The embedding $ e_{j} $ for each input token $ x_{j} $ with factors $ F $ is created as in  Eq.~\ref{eq:factored_input}:

\begin{equation}
    e_{j} = \bigparallel_{k=1}^{|F|} \mathbf{E}_{k} x_{jk} 
    \label{eq:factored_input}
\end{equation}

\noindent where $ \bigparallel $ indicates vector concatenation, $ \mathbf{E}_{k} $ is the embedding matrix of factor $ k $, and $ x_{jk} $ is a one hot vector for the $k$-th input factor. Table \ref{tab:features} lists each of the factors used in our models, along with its corresponding embedding size. The embedding size of 300 for the lemma is set in configuration, while the embedding sizes of the other features are set heuristically by OpenNMT-py, using the heuristic $ |embedding_{k}| = |V_{k}|^{0.7} $, where $ |V_{k}| $ is the vocabulary size of feature $ k $. Table \ref{tab:input-data-example} gives an example from the training data with actual instantiations of each of the features. 

\section{Model}

Models were trained using the OpenNMT-py toolkit \citep{Klein2017}. The model architecture is a 1 layer bidirectional recurrent neural network (RNN) with long short-term memory (LSTM) cells \citep{Hochreiter1997} and attention \citep{Luong2015EffectiveTranslation}. The model has 450 hidden units in the encoder and decoder layers, and 300 hidden units in the word embeddings which are learned jointly across the whole model. Dropout of 0.3 is applied between the LSTM stacks. We use a coverage attention layer \citep{Tu2016ModelingTranslation} with lambda value of 1.

The models are trained using stochastic gradient descent with learning rate 1. A learning rate decay of 0.5 is applied at each epoch once perplexity does not decrease on the validation set. Models were trained for 20 epochs. Output was decoded using beam search with beam size 5. Unknown tokens were replaced with the input token that had the highest attention value at that time step \citep{Vinyals2015}. Output from the epoch checkpoint which performed best on the development set was chosen for test set submission. 

The exploration and choice of hyperparameters was aided by the use of Bayesian hyperparameter optimization platform SigOpt\footnote{\url{https://sigopt.com/}}. 






\section{Experiments}


We experiment with many different combinations of input features and training data, in order to understand which elements of the representation have the largest impact upon performance. 

We limit vocabulary size during training to enable the pointer network to generalize to unknown tokens at test time. When using just the SR training data we train word embeddings for the 15,000 most frequent tokens from a possible 23,650 unique tokens. When using the combined SR training data and filtered WikiText dataset we use the 30,000 most frequent tokens from a possible 106,367 unique tokens.

We trained on a single Tesla K40 GPU. Training time was approximately 1 minute per epoch for the SR data and 1 hour per epoch for the combined SR data and filtered WikiText.

\section{Results}

We report results using automated evaluation metric BLEU \citep{Papineni:2002:BMA:1073083.1073135}. On the test set we additionally report the NIST \citep{Przybocki2009} score and the normalized edit distance (DIST).

\begin{table}[!ht]
\centering
\begin{tabular}{p{0.6\linewidth}p{0.2\linewidth}}
\toprule
\textsc{System} & \textsc{BLEU} \\ \midrule
SR Baseline & 21.27 \\
SR + delemma suggestions & 23.75 \\
SR + delemma suggestions + linearization & 43.11 \\
SR + delemma suggestions + linearization + additional data & 68.86 \\
\bottomrule
\end{tabular}
\caption{Ablation study with BLEU scores for different configurations on the shallow task development set}
\label{tab:ablation-results}
\end{table}

Table \ref{tab:ablation-results} presents the results of the surface realization experiments. We observe three main components that drastically improve performance over the baseline model:

\begin{enumerate}
    \item augmenting the training set with more data
    \item reordering the input using the dependency parse features
    \item providing potential forms via the delemmatization map
\end{enumerate}

Table \ref{tab:srst-official-results} gives the official SR 2018 results from task organizers. Our system, which corresponds to the best configuration from Table \ref{tab:ablation-results} was ranked first across all metrics. 

\begin{table}[!ht]
\centering
\begin{tabular}{lccc}
\toprule
\textsc{Team ID} & \textsc{BLEU} & \textsc{DIST} & \textsc{NIST} \\ \midrule
1 (Ours) & \textbf{69.14} & \textbf{80.42} & \textbf{12.02} \\
2 & 28.09 & 70.01 & 9.51 \\ 
3 & 8.04 & 47.63 &  7.71 \\
4 & 66.33 & 70.22 & \textbf{12.02} \\
5 & 50.74 & 77.56 & 10.62 \\
6 & 55.29 & 79.29 & 10.86 \\
7 & 23.2 & 51.87  & 8.86 \\
8 & 29.6 & 65.9  & 9.58  \\
\midrule
AVG & 41.3 & 67.86 & 10.15 \\
\bottomrule
\end{tabular}
\caption{Official results of the surface realization shared task using BLEU, DIST and NIST as evaluation metrics.}
\label{tab:srst-official-results}
\end{table}

\section{Related Work}

The surface realization task bears the closest resemblance to the SemEval 2017 shared task AMR-to-text \citep{May2017}. Our approach to data augmentation and preprocessing uses many insights from Neural AMR \citep{Konstas2017NeuralGeneration}. Traditional data-to-text systems use a rule based approach \citep{Reiter:2000:BNL:331955}.




\section{Conclusion}

The main takeaway from this work is that data augmentation improves performance on the surface realization task. Although unsurprising, this result confirms that sufficient data is needed to achieve reasonable performance, and that flattened structural information such as dependency parse features is insufficient without additional preprocessing to reduce the complexity of the input. The surface realization task is ostensibly quite simple, thus it is surprising that baseline sequence to sequence models, which perform well in other tasks such as machine translation, cannot solve this task. We hypothesize that the lemmatization and shuffling of the input does not provide sufficient information to reconstruct the input. In sequences longer than a few words, there is likely to be significant ambiguity without additional structural information such as parse features. However, reconstructing the original sequence from unprocessed, flattened parse information alone is unrealistic using standard encoder-decoder models. 

In future work, we plan to explore more challenging variants of this task, while also experimenting with models that do not require feature-specific preprocessing to make use of rich structural information in the input.




\bibliography{mendeley_v2,acl2018}
\bibliographystyle{acl_natbib}

\end{document}